\documentclass[runningheads,a4paper]{llncs}
\usepackage{dsfont}
\usepackage{amsmath}
\usepackage{amssymb}
\usepackage{graphicx}
\usepackage{subfigure}
\usepackage{wrapfig}
\usepackage{cite}
\usepackage{bm}
\usepackage{hyperref}
\usepackage{url}

\usepackage{tabularx}
\usepackage{booktabs}

\begin{document}

\mainmatter  

\title{Semi-Supervised Brain Lesion Segmentation with an Adapted Mean Teacher Model}

\titlerunning{Semi-Supervised Brain Lesion Segmentation}

%
%
\author{Wenhui Cui$^{1}$, Yanlin Liu$^{2}$, Yuxing Li$^{2}$, Menghao Guo$^{1}$, Yiming Li$^{3}$, Xiuli Li$^{3,4}$, Tianle Wang$^{5}$, Xiangzhu Zeng$^{6}$, Chuyang Ye$^{2}$}
\authorrunning{Cui et al.}

\institute{$^{1}$School of Computer Science and Technology, Xidian University, Xi'an, China\\
$^{2}$School of Information and Electronics, Beijing Institute of Technology, Beijing, China\\
$^{3}$Deepwise AI Lab, Beijing, China\\
$^{4}$Peng Cheng Laboratory, Shenzhen, China\\
$^{5}$Department of Radiology, The People's Hospital of Nantong, Nantong, China\\
$^{6}$Department of Radiology, Peking University Third Hospital, Beijing, China}

%
%

\toctitle{}
\tocauthor{}
\maketitle

\begin{abstract}

Automated brain lesion segmentation provides valuable information for the analysis and intervention of patients. In particular, methods that are based on \textit{convolutional neural networks}~(CNNs) have achieved state-of-the-art segmentation performance. However, CNNs usually require a decent amount of annotated data, which may be costly and time-consuming to obtain. 
Since unannotated data is generally abundant, it is desirable to use unannotated data to improve the segmentation performance for CNNs when limited annotated data is available. 
In this work, we propose a \textit{semi-supervised learning}~(SSL) approach to brain lesion segmentation, where unannotated data is incorporated into the training of CNNs. 
We adapt the mean teacher model, which is originally developed for SSL-based image classification, for brain lesion segmentation.
Assuming that the network should produce consistent outputs for similar inputs, a loss of segmentation consistency is designed and integrated into a self-ensembling framework.
Self-ensembling exploits the information in the intermediate training steps, and the ensemble prediction based on the information can be closer to the correct result than the single latest model.
To exploit such information, we build a student model and a teacher model, which share the same CNN architecture for segmentation.
The student and teacher models are updated alternately. At each step, the student model learns from the teacher model by minimizing the weighted sum of the segmentation loss computed from annotated data and the segmentation consistency loss between the teacher and student models computed from unannotated data. 
Then, the teacher model is updated by combining the updated student model with the historical information of teacher models using an exponential moving average strategy.
For demonstration, the proposed approach was evaluated on ischemic stroke lesion segmentation.
Results indicate that the proposed method improves stroke lesion segmentation with the incorporation of unannotated data and outperforms competing SSL-based methods.

\end{abstract}

\section{Introduction}
\label{sec:intro}

Automated segmentation of brain lesions in \textit{magnetic resonance images}~(MRIs) provides valuable information for the analysis and intervention of patients~\cite{kamnitsas2017efficient}. 
Deep learning based approaches have been developed for the segmentation of different types of brain lesions, such as stroke lesions~\cite{kamnitsas2017efficient,kuang2018joint} and brain tumors~\cite{zhao2018deep,pereira2016brain,havaei2017brain}. 
Various architectures of \textit{convolutional neural networks}~(CNNs) have been proposed and have achieved state-of-the-art segmentation performance.
Deep learning based approaches usually involve a huge number of parameters and thus require a decent amount of annotated data, so that the parameters can be properly learned~\cite{zhang2017deep}. However, manual annotation of brain lesions is costly and time-consuming, whereas unannotated data is often abundant.
Therefore, it is desirable to exploit the unannotated data when there is limited annotated data for training. 

\textit{Semi-supervised learning}~(SSL) techniques have emerged as means to combine the limited annotated data and the abundant unannotated data to improve the training process~\cite{zhou2017brief}. Several methods have been proposed for medical image segmentation~\cite{zhang2017deep,baur2017semi}. For example, the consistency of feature embedding between annotated and unannotated data is enforced in~\cite{baur2017semi}, where a consistency loss is incorporated into the loss function and provides regularization for training the CNN. A similar idea is developed in~\cite{kamnitsas2017unsupervised}, where the consistency of feature embedding is ensured with an adversarial learning strategy. Note that although the approach in~\cite{kamnitsas2017unsupervised} is originally developed for transfer learning, it can be applied to SSL as well. Another approach in~\cite{zhang2017deep} aims to achieve similar quality of segmentation on the annotated and unannotated data. The similarity is encouraged with a deep adversarial network model, which consists of a segmentation network and an evaluation network. These approaches have achieved promising results when limited annotated data is available. However, the development of SSL techniques for CNN-based brain lesion segmentation is still an open problem, where improved segmentation performance is desired.

In this work, we explore the integration of SSL into CNN-based brain lesion segmentation. Inspired by the success of the \textit{mean teacher}~(MT) model~\cite{tarvainen2017mean} for SSL-based image classification, we propose an adapted MT model for brain lesion segmentation, where both annotated and unannotated data can be exploited to boost segmentation performance. 

We assume that the segmentation should be consistent for similar input data~\cite{laine2016temporal}, and define a segmentation consistency loss, which is computed for a pair of inputs that are obtained by adding noises to the same unannotated sample. 
In this way, unannotated data can be incorporated into the learning process and provide regularization information.
Note that unlike in previous works~\cite{baur2017semi,zhang2017deep} that measure the consistency between annotated data and unannotated data, here the consistency is computed between two noisy versions of the same unannotated data.
Since it is observed in~\cite{laine2016temporal} and \cite{tarvainen2017mean} that self-ensembling could lead to better classification models, we apply a similar strategy to brain lesion segmentation and integrate the segmentation consistency loss into the self-ensembling framework.
Specifically, we build a teacher model and a student model, which share the same network architecture.
In this work we select the DeepMedic architecture~\cite{kamnitsas2017efficient} for the two models, because it has achieved state-of-the-art performance in brain lesion segmentation~\cite{maier2017isles}.
Self-ensembling is based on the observation that the ensemble prediction combining the network information after each step is more accurate than the current output~\cite{laine2016temporal,tarvainen2017mean}.
Thus, the teacher model records the information at each step, and the student model learns from the teacher model by minimizing the loss of segmentation accuracy for annotated data and the consistency loss with respect to the outputs of the teacher model for unannotated data.
Then, the teacher model is updated by combining the historical information of teacher models and the current student model with an \textit{exponential moving average}~(EMA) strategy. 
The student and teacher models are updated alternately, and the final teacher model is used for segmentation on test samples.

For demonstration, the proposed approach was evaluated on ischemic stroke lesion segmentation. Results indicate that the proposed method improves the segmentation quality by incorporating unannotated data and outperforms competing SSL-based segmentation strategies.

\section{Methods}
\label{sec:method}
In this section, we first introduce the backbone CNN architecture shared by the teacher and student models.
Then, we describe how unannotated data is used by the teacher and student models to regularize the model training.
Finally, implementation details are given.

\subsection{Backbone CNN Architecture}

\begin{figure}[!t]
  \centering
	\includegraphics[width=1.0\columnwidth]{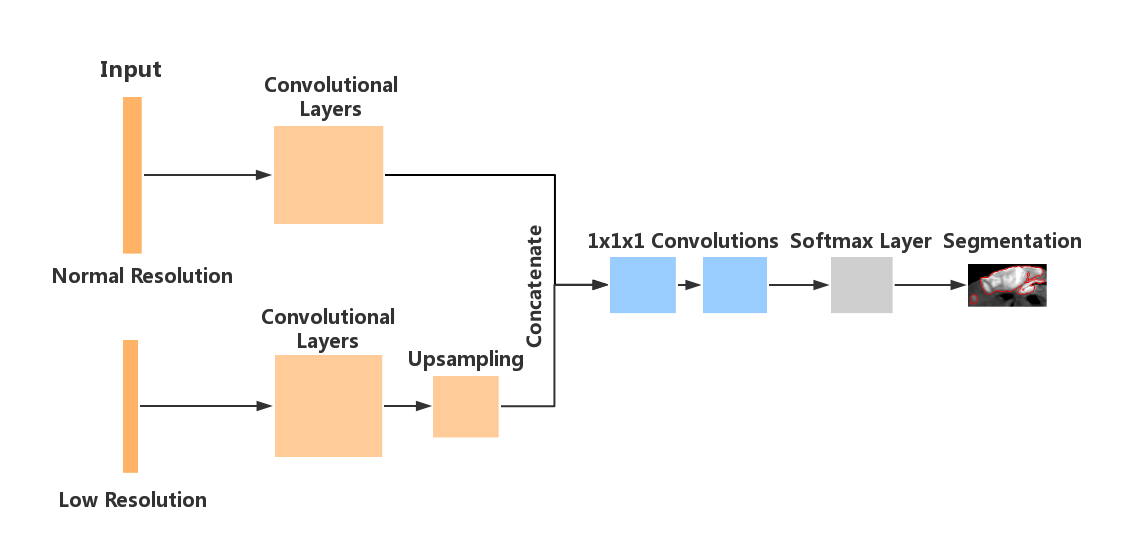}
\caption{The DeepMedic model proposed in~\cite{kamnitsas2017efficient}.}
\label{fig:deepmedic}
\end{figure}

Due to its superior segmentation performance, the DeepMedic model~\cite{kamnitsas2017efficient} is used as our backbone network structure, which is shared by the teacher and student models. Specifically, DeepMedic is a dual pathway, 11-layer deep, three-dimensional CNN, and it performs multi-scale processing via parallel convolutional pathways. 
A graphical illustration of DeepMedic is shown in Fig.~\ref{fig:deepmedic}, and the parameters of each layer are summarized in Table~\ref{tab:para}. 
Note that the two pathways use the same settings of convolutional layers 1-8 in Table~\ref{tab:para}. 
DeepMedic takes image patches at two different resolutions as input. The two patches are centered at the same image location. The upper pathway in~Fig.~\ref{fig:deepmedic} takes normal resolution image patches as input, whereas the bottom pathway in~Fig.~\ref{fig:deepmedic} operates on downsampled patches (by a factor of three). 
Before the final segmentation, the multi-scale features are concatenated and fed into $1^{3}$ convolutional layers. 
For more details about DeepMedic, we refer readers to~\cite{kamnitsas2017efficient}.

\begin{table}[!t]
\centering
\caption{The specification of layers in DeepMedic.}
\begin{tabular}{p{4.5cm}<{\centering}p{4.5cm}<{\centering}}
\textbf{Layer} & \textbf{Parameters} \\
\toprule[0.8pt]
Convolutional layer 1& 30 filters, $3^3$\\ 
Convolutional layer 2& 30 filters, $3^3$\\ 
Convolutional layer 3& 40 filters, $3^3$\\ 
Convolutional layer 4& 40 filters, $3^3$\\ 
Convolutional layer 5& 40 filters, $3^3$\\ 
Convolutional layer 6& 40 filters, $3^3$\\ 
Convolutional layer 7& 50 filters, $3^3$\\ 
Convolutional layer 8& 50 filters, $3^3$\\ 
$1^3$ Convolution & 250 filters, $1^3$\\
$1^3$ Convolution & 250 filters, $1^3$\\
\label{tab:para}
\end{tabular}
\end{table}

\subsection{Semi-Supervised Lesion Segmentation with an Adapted MT Model}
\label{sec:mt}

To leverage the abundant unannotated data for lesion segmentation, we propose to use an SSL strategy. 
Our strategy is inspired by the MT model~\cite{tarvainen2017mean}, which is developed for SSL-based image classification. 
Like in the MT model, we assume that CNN models should favor functions that produce consistent outputs for similar inputs.
Pairs of similar input samples are generated by adding noises to the same unannotated data. In this way, unannotated data can be used to provide regularization for training the network.
Note that unlike MT, we need to measure the consistency of segmentation instead of classification. Thus, we adapt the MT strategy by defining a segmentation consistency loss.
The segmentation consistency loss is then integrated into a self-ensembling framework, which is motivated by the observation that the ensemble prediction based on the combined information after each step can be more accurate than the current output~\cite{tarvainen2017mean}.
The detailed description of the proposed approach is given below.

For an unannotated input $X_{\mathrm{u}}$, we add noises $\eta$ and $\eta'$ sampled from the same distribution, and the network is expected to produce similar outputs for the two noisy inputs. Although it is possible to directly incorporate a consistency loss based on the similarity into DeepMedic, which leads to a strategy similar to the $\Pi$ model in~\cite{laine2016temporal} for classification, integration of the consistency loss into a self-ensembling framework can lead to better model training~\cite{tarvainen2017mean}.
Therefore, like the original MT approach, we build a teacher model and a student model, where the student model attempts to learn the targets generated by the teacher model.

Both the teacher and student models share the same DeepMedic architecture~\cite{kamnitsas2017efficient}. 
Note that the proposed framework is not restricted to a specific segmentation network, and can be applied to other backbone segmentation architectures as well, such as 3D U-Net~\cite{cicek20163d}. 
The two noisy inputs associated with $\eta$ and $\eta'$ are then fed into the student model and the teacher model, respectively.
Since the student and teacher models share the same architecture, we denote their output for the noisy input as $f(X_{\mathrm{u}},\eta,\theta)$ and $f(X_{\mathrm{u}},\eta',\theta')$, respectively. Here, $\theta$ and $\theta'$ are the weights in the network of the student model and the teacher model, respectively.

The teacher model is initialized with the DeepMedic network trained with annotated data.
Then, the teacher and student models are updated alternately.
At each step, the student model learns from the teacher model by minimizing the weighted sum of the consistency loss $\mathcal{L}_{\mathrm{c}}$ of unannotated data and segmentation loss $\mathcal{L}_{\mathrm{s}}$ of annotated data. Specifically, we define $\mathcal{L}_{\mathrm{c}}$ as the soft Dice loss between the predicted probability maps of the student and teacher models based on their corresponding noisy inputs 
\begin{eqnarray}
\mathcal{L}_{\mathrm{c}} = 1 - \mathbb{E}_{X_{\mathrm{u}},\eta,\eta'}\left[\frac{1}{K} \sum_{i=1}^{K}\frac{\sum_{v=1}^{V} 2f^{i}_{v}(X_{\mathrm{u}},\eta,\theta) f^{i}_{v}(X_{\mathrm{u}},\eta',\theta')} {\sum_{v=1}^{V} f^{i}_{v}(X_{\mathrm{u}},\eta,\theta) + \sum_{v=1}^{V} f^{i}_{v}(X_{\mathrm{u}},\eta',\theta')}\right]
\label{eq:consistency}
\end{eqnarray}
where $f^{i}_{v}(\cdot)$ represents the $f(\cdot)$ value that is at the $v$-th voxel and takes the $i$-th label, $K$ represents the total number of possible labels, and $V$ denotes the total number of voxels in the input.
With the loss defined in Eq.~(\ref{eq:consistency}), the output of the teacher model can also be considered a target label for the student model to learn.
As in DeepMedic~\cite{kamnitsas2017efficient}, $\mathcal{L}_{\mathrm{s}}$ is the cross entropy loss between the predictions $f(X_{\mathrm{a}},\theta)$ (no noise $\eta$ is applied) of the student model for the annotated input $X_{\mathrm{a}}$ and the corresponding annotation $Y$
\begin{eqnarray}
\mathcal{L}_{\mathrm{s}} = - \mathbb{E}_{X_{\mathrm{a}},Y}\left[ \frac{1}{V}\sum_{v=1} ^{V}\sum_{i=1} ^{K}  Y^{i}_{v} \log\left(f^{i}_{v}(X_{\mathrm{a}},\theta)\right)\right],	
\label{eq:seg}
\end{eqnarray}
where $Y^{i}_{v}$ represents the value of $Y$ at the $v$-th voxel with label $i$.

Then, the total loss function $\mathcal{L}$ of our model is
\begin{eqnarray}
\mathcal{L} = \mathcal{L}_{\mathrm{s}} + \beta \mathcal{L}_{\mathrm{c}},
\label{eq:loss}
\end{eqnarray}
where $\beta$ is an adaptive weighting coefficient. As suggested in~\cite{tarvainen2017mean}, different $\beta$ is used at different steps.
Specifically, $\beta=\exp\left(-5(1 - \frac {S}{L}) ^2 \right)$ (when $S\leq L$), where $S$ is the current training step and $L$ is called the ramp-up length; when $S> L$, $\beta$ is set to one. In our experiment, we empirically set $L = 400$.
Such an adaptive setting of $\beta$ keeps the effect of consistency down in early steps, because the teacher model may not generate reasonable target labels at the beginning~\cite{tarvainen2017mean}. 

With the parameters $\theta_{t}$ of the student model at step $t$, we perform the EMA of weights to aggregate information in training steps as in the original MT model~\cite{tarvainen2017mean}. Specifically, we update the teacher model as follows
\begin{eqnarray}
\theta'_{t} = \alpha \theta'_{t-1} + (1-\alpha) \theta_{t},
\label{eq:ema}
\end{eqnarray}
where $\alpha$ is the EMA decay.
Compared with other ensembling strategies~\cite{laine2016temporal}, EMA better prevents overfitting, especially when a large number of model parameters are learned from limited training data~\cite{tarvainen2017mean}.  
Following the MT method~\cite{tarvainen2017mean}, we used $\alpha =0.99$ in the first $L$ steps (the ramp-up phase), and $\alpha =0.999$ for the rest of the training. This strategy facilitates the teacher model to 1) forget the old inaccurate student weights quickly and 2) benefit from a longer memory when the student improvement slows down after the ramp-up phase.
The final teacher model is used to perform lesion segmentation for test samples.

\subsection{Implementation Details}

The proposed method is implemented using TensorFlow (\url{https://www.tensorflow.org}). 
In the training of student models, we followed the settings in~\cite{kamnitsas2017efficient} and minimized the loss with an RMSProp optimizer~\cite{tieleman2012lecture}, where the learning rate is 0.0001 and the decay rate is 0.9. The batch size is 16, which consists of eight annotated and eight unannotated samples.
Both the annotated and unannotated training patches were sampled from the lesion region and healthy tissue with equal probability, which mitigates class imbalance~\cite{kamnitsas2017efficient}.
Note that since the lesion region is unknown for unannotated data, it is approximated by the DeepMedic prediction.

The noise injection for the proposed method was applied as follows.
We applied Gaussian noises to the inputs of the student and teacher models. Noise~$\eta$ consists of two different types: the additive noise $\eta_{\mathrm{s}}$ and the multiplicative noise $\eta_{\mathrm{m}}$. Both are sampled from Gaussian distributions. At each voxel of the input patch, noise was applied independently, and the noisy intensity $I'$ is computed from the original intensity $I$ as follows
\begin{eqnarray}
I' = (I + \eta_\mathrm{s})\times \eta_\mathrm{m}.
\end{eqnarray}

\section{Experiments}

\subsection{Data Description}

For demonstration, the proposed method was evaluated on a task of ischemic stroke lesion segmentation. A total number of 246 \textit{diffusion weighted images}~(DWIs) of ischemic stroke patients were acquired on a 3T Siemens Verio scanner, where a $b$-value of $1000~\mathrm{s}/\mathrm{mm}^{2}$ was applied and a $b0$ image (the image without diffusion weighting) was also acquired. The image resolution is $0.96~\mathrm{mm}\times 0.96~\mathrm{mm}\times 6.5~\mathrm{mm}$ and the image dimension is $240\times 240 \times 21$. Manual delineations of stroke lesions were performed by an experienced radiologist on 50 DWIs, and the rest 196 DWIs are unannotated. 

The image intensities were normalized for each scan. Specifically, a brain mask was extracted with the Dipy software~\cite{garyfallidis2014dipy}.
Then, the mean and standard deviation of the intensity in the brain were computed. The mean was subtracted from the intensity at each voxel in the skull-stripped image and the resulting intensities were further divided by the standard deviation.
The patch sizes for the normal resolution and downsampled pathways are $37\times 37\times21$ and $23\times 23\times 18$, respectively, so that the multi-scale features can be concatenated.
The additive noise was sampled from a Gaussian distribution which has a zero mean and a standard deviation of 0.05, whereas the multiplicative noise was sampled from a Gaussian distribution which has a mean of 1.0 and a standard deviation of 0.01.

\subsection{Training Phase}
\begin{figure}[!t]
  \centering
	\includegraphics[width=0.85\columnwidth]{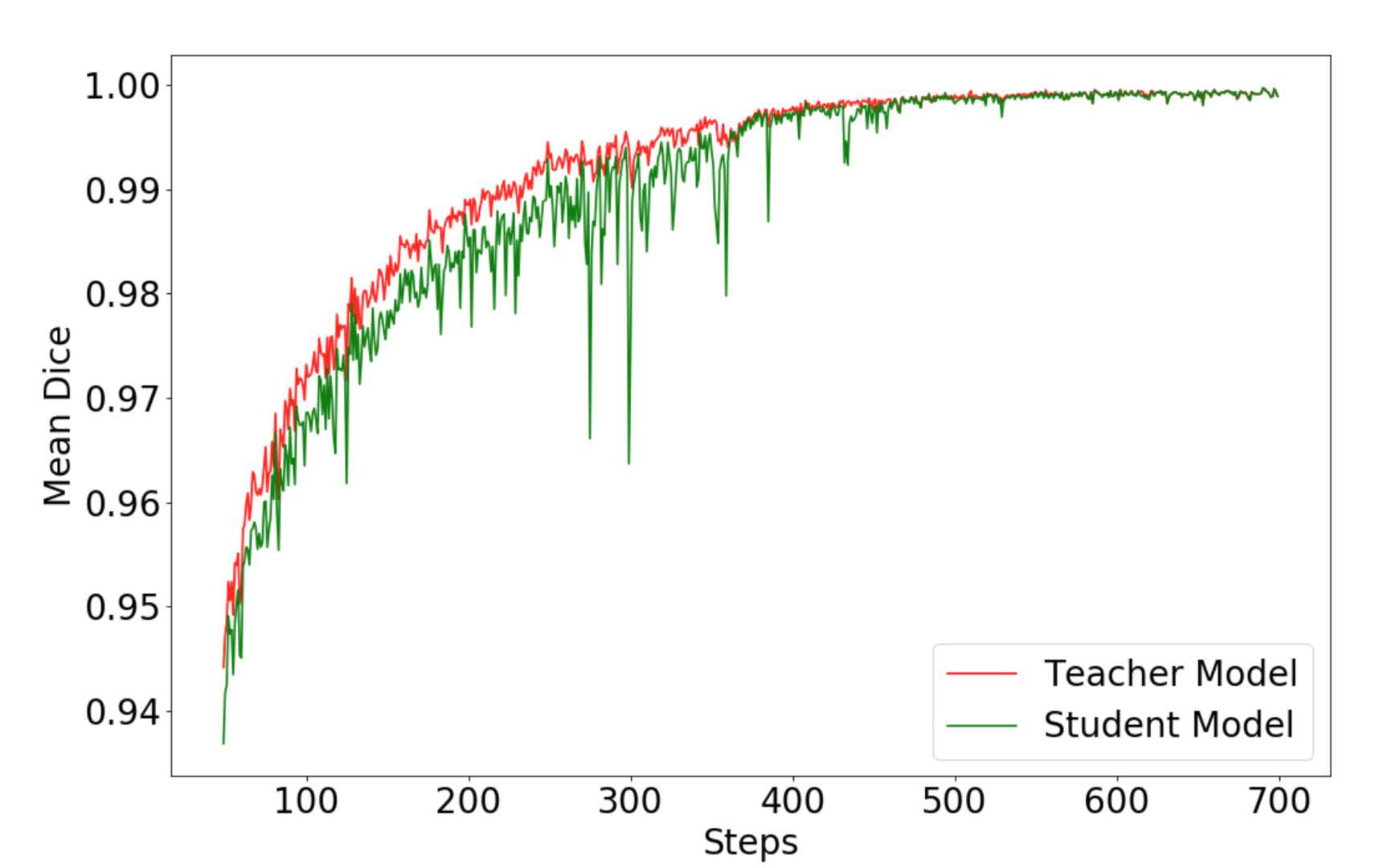}
\caption{The mean Dice coefficients of the training data for the teacher model and the student model after each training step.}
\label{fig:training}
\end{figure}

We randomly selected 20 annotated subjects as training scans, and used the rest 30 annotated data as test scans. The 196 unannotated scans were included in the training process of the proposed approach as well. 
The training was performed on an NVIDIA GeForce GTX 1080Ti GPU, and it took about 12 hours.
The training process was evaluated and the Dice coefficients of the training data are shown for the student and teacher models in Fig.~\ref{fig:training}. We can see that both models better fit the training data as the training continues and become stable in the end. In addition, the teacher model consistently achieves higher Dice coefficients than the student model, until it is close to the end of the ramp-up phase (400 steps), where the two Dice coefficients are close.
These observations are consistent with the assumption in self-ensembling and the settings of the EMA decay.

\subsection{Evaluation of Lesion Segmentation}

\begin{figure}[!t]
  \centering
	\includegraphics[width=0.8\columnwidth]{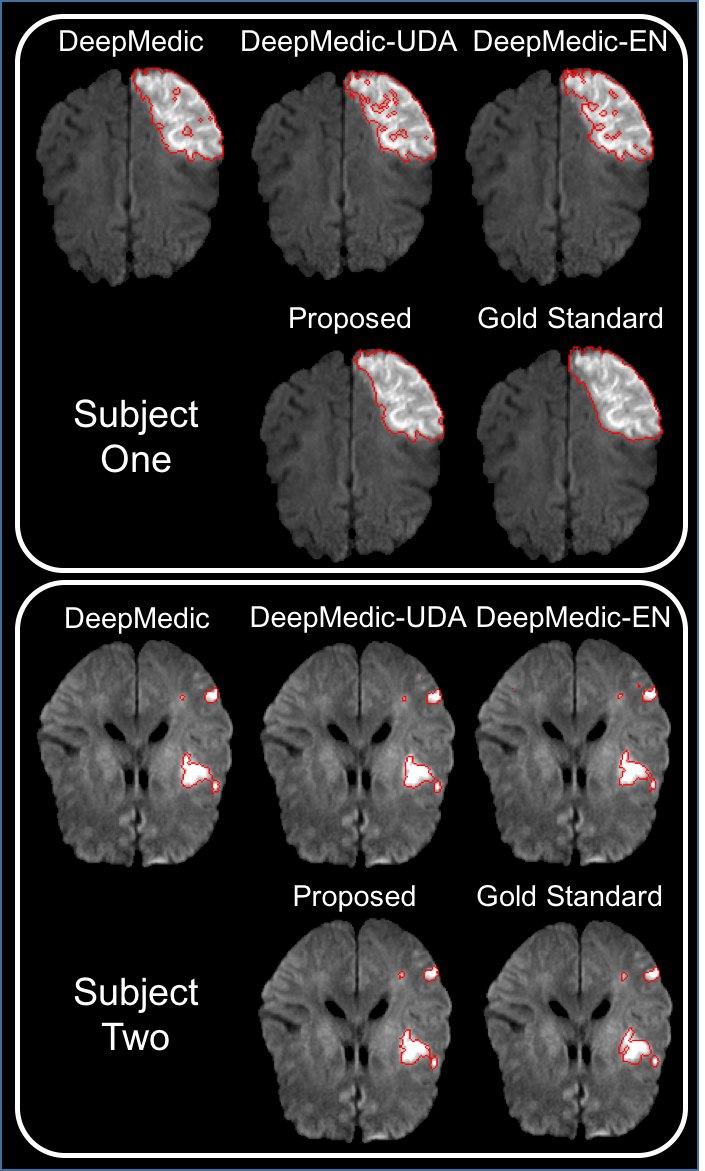}
\caption{Cross-sectional views of the segmentation results overlaid on DWIs for two representative test subjects with different sizes of lesions.}
\label{fig:results}
\end{figure}

Then, we evaluated the segmentation results of the proposed method. The proposed method was compared with three methods. The DeepMedic approach~\cite{kamnitsas2017efficient} was included as the baseline method that does not use unannotated data for training. The strategy used by~\cite{zhang2017deep} was also integrated with DeepMedic for comparison, which performs SSL-based image segmentation. This strategy assumes that the segmentation of unannotated data should follow a distribution that is similar to that of annotated data. Such similarity is enforced by a separate evaluation network with adversarial learning. We replaced the segmentation network in~\cite{zhang2017deep} with DeepMedic for lesion segmentation. Due to the use of an evaluation network, this strategy is referred to as DeepMedic-EN. The network in~\cite{kamnitsas2017unsupervised} for unsupervised domain adaptation was also considered, because although it is originally developed for transfer learning, it can be directly used for SSL. This strategy applies an idea that is similar to~\cite{zhang2017deep}, where adversarial learning is applied so that the features extracted from the source (annotated) and target (unannotated) data follow similar distributions. Since the deep network in~\cite{kamnitsas2017unsupervised} is based on the DeepMedic architecture, we used the structure directly, and the method is referred to as DeepMedic-UDA, where UDA stands for unsupervised domain adaptation as described by~\cite{kamnitsas2017unsupervised}.

We first qualitatively evaluated the proposed method. Cross-sectional views of the segmentation results overlaid on DWIs are shown in Fig.~\ref{fig:results} for two representative test subjects with different sizes of lesions. The gold standard of the manual delineation and the results of the competing methods are also shown for comparison. It can be seen that the proposed method produced segmentation that better agrees with the gold standard.

Next, the proposed method was quantitatively evaluated. We computed the Dice coefficients on the test scans for the proposed and competing methods, and the results are summarized in Table~\ref{tab:results}. Here, the means and standard deviations of the Dice coefficients computed from the 30 test subjects are listed. The proposed method has the highest mean Dice coefficients, which indicates its better segmentation quality than the competing methods. In addition, the results were compared between the proposed method and each competing method with a paired Student's $t$-test. In all cases, the difference is significant ($p<0.05$).
Note that DeepMedic-EN and DeepMedic-UDA have smaller mean Dice coefficients than the baseline DeepMedic. This is possibly due to the limited number of training scans, which cannot adequately represent the distribution of annotated data. Thus, the adversarial learning in DeepMedic-EN and DeepMedic-UDA may incorrectly modify the segmentation result.

\begin{table}[!t]
\centering
\caption{Means and standard deviations of the Dice coefficients on test scans when 20 annotated scans were used for training. Best results are highlighted in bold font. Asterisks ($^{*}$) indicate that the difference between the proposed method and the competing method is statistically significant ($p<0.05$) using a paired Student's $t$-test.}
\begin{tabular}{p{3cm}<{\centering}p{3cm}<{\centering}p{3cm}<{\centering}p{3cm}<{\centering}}
\toprule[1.3pt]
      DeepMedic  & DeepMedic-UDA   & DeepMedic-EN & Proposed\\
\midrule[0.8pt]

$0.6312 \pm 0.2617^{*}$ & $0.6096 \pm 0.2819^{*}$   &   $0.6236 \pm 0.2577 ^{*}$ & $\mathbf{0.6676 \pm 0.2392}$ \\

\bottomrule[1.3pt]
\label{tab:results}
\end{tabular}
\end{table}

\begin{table}[!t]
\centering
\caption{Means and standard deviations of the Dice coefficients on test scans when 10 and 30 annotated scans were used for training. Best results are highlighted in bold font. Asterisks ($^{*}$) indicate that the difference between the proposed method and the competing method is statistically significant ($p<0.05$) using a paired Student's $t$-test.}
\begin{tabular}{p{2.5cm}<{\centering}p{4.5cm}<{\centering}p{4.5cm}<{\centering}}

\toprule[1.3pt]
      & 10 Annotated Training Scans \quad & 30 Annotated Training Scans\\
\midrule[0.8pt]
DeepMedic  & $ 0.5912 \pm  0.2243^{*}$ &  $0.6551 \pm 0.2514^{*}$\\

DeepMedic-UDA & $0.5562\pm0.2878^{*} $ & $ 0.6627 \pm 0.2208^{*} $  \\  

DeepMedic-EN &  $0.5684 \pm 0.2873^{*}$  & $0.6503 \pm 0.2650^{*}$\\   

Proposed &  $\mathbf{0.6518 \pm 0.2484}$ &  $\mathbf{0.6879 \pm 0.2334}$\\
\bottomrule[1.3pt]
\end{tabular}
\label{tab:results2}
\end{table}

\subsection{Impact of the Amount of Training Data}

Lastly, we investigated the impact of the number of training scans. Specifically, we investigated two additional cases, where 10 and 30 randomly selected annotated scans were included in training and the rest annotated data were used for testing. For SSL-based methods, all the unannotated data were also included in training. The results are shown in Table~\ref{tab:results2}, where the means and standard deviations of the Dice coefficients are listed. In all cases, the proposed approach has higher mean Dice coefficients than the competing methods, and the difference is significant using a paired Student's $t$-test. These results indicate that the proposed method outperforms the competitors.

\section{Discussion}

The original MT model is developed for semi-supervised image classification, and its consistency loss is simply the difference between class predictions. In our task, however, the consistency needs to be enforced for segmentation. Thus, we have adapted the MT model by defining the consistency loss based on the Dice coefficient. The results indicate that the adaption can be successfully applied to semi-supervised image segmentation.

We have performed experiments with different numbers of training scans. As expected, a greater number of training scans leads to more accurate segmentation for all the methods considered in the experiment. In addition, when the number of training scans is small, the SSL-based approaches DeepMedic-EN and DeepMedic-UDA perform even worse than the baseline DeepMedic model. It is possibly due to the small number of training scans, which cannot adequately represent the distribution of desired features and segmentation. Thus, the adversarial learning strategy in DeepMedic-EN and DeepMedic-UDA cannot enforce proper regularization based on the unannotated data, where it is very likely that the segmentation or the extracted feature of unannotated data does not resemble that of the annotated data. As the number of training scans increases, the margin between the baseline DeepMedic and DeepMedic-EN or DeepMedic-UDA becomes smaller, possibly because the annotated data can better represent the distribution of expected features and segmentation. With 30 training scans, DeepMedic-UDA is able to outperform the baseline DeepMedic.

Unlike DeepMedic-EN and DeepMedic-UDA, the proposed approach relies on the assumption that similar inputs should produce consistent outputs, and the use of such consistency is further improved with a self-ensembling strategy. In contrast to the adversarial learning strategies in DeepMedic-EN and DeepMedic-UDA, the adapted MT model in the proposed work is less affected by the limited number of training scans, because it does not require the comparison between annotated data and unannotated data. This is confirmed by the results, where the propose approach is robust to the decrease in the number of training scans.

We have also observed that with only 10 annotated training scans, the proposed method outperforms the baseline DeepMedic model trained by 20 annotated scans and performs comparably to the baseline DeepMedic model trained by 30 annotated scans. This highlights the importance of the incorporation of unannotated data for training CNNs. It can potentially greatly reduce the annotation cost or increase the segmentation quality with existing annotated data.

We applied Gaussian noise to the input samples to generate a pair of similar inputs for the student and teacher models.
Other strategies for generating pairs of similar inputs are possible. For example, dropout provides a convenient way for noise injection~\cite{wager2013dropout}.
Future works may explore additional approaches to enforcing the consistency regularization to more efficiently use unannotated data.

\section{Conclusion}
\label{sec:conclusion}

We have proposed an SSL-based approach to brain lesion segmentation. A teacher model and a student model are constructed and updated alternately. By minimizing the segmentation loss computed from annotated data and segmentation consistency loss computed from unannotated data, the student model learns from the teacher model at each step. The teacher model is then updated with an EMA strategy, and the final teacher model performs lesion segmentation on test samples. The proposed method was applied to ischemic stroke lesion segmentation, and the results demonstrate the benefit of incorporating unannotated data using the proposed method.

\section*{Acknowledgement}
This work is supported by the National Natural Science Foundation of China (61601461), Beijing Natural Science Foundation~(7192108), and Beijing Institute of Technology Research Fund Program for Young Scholars. 

\bibliographystyle{splncs03}
\bibliography{refs}
\end{document}